\documentclass[conference]{IEEEtran}
\IEEEoverridecommandlockouts
\usepackage{cite}
\usepackage{amsmath,amssymb,amsfonts}
\usepackage{algorithmic}
\usepackage{graphicx}
\usepackage{textcomp}
\usepackage{xcolor}

\usepackage{adjustbox}
\usepackage{diagbox}
\usepackage{array}

\usepackage{xcolor,colortbl}
\usepackage[nomessages]{fp} 

\usepackage{makecell}

\newcolumntype{R}[2]{%
    >{\adjustbox{angle=#1,lap=\width-(#2)}\bgroup}%
    r%
    <{\egroup}%
}
\newcommand*\rot{\multicolumn{1}{|R{90}{2em}|}}

\def\BibTeX{{\rm B\kern-.05em{\sc i\kern-.025em b}\kern-.08em
    T\kern-.1667em\lower.7ex\hbox{E}\kern-.125emX}}
    
\usepackage{enumitem}

\newcounter{SentenceCounter}
\newcounter{ExampleCounter}

\newenvironment{Sentence}[1][]{
\refstepcounter{SentenceCounter}
\vspace{0.5mm}
\def\temp{#1}\ifx\temp\empty
    \noindent Sentence~\theExampleCounter.\theSentenceCounter:
\else
    \noindent Sentence~\theExampleCounter.\theSentenceCounter\hspace{1mm}(#1):
\fi
\itshape
}
{
\label{sent\theExampleCounter.\theSentenceCounter}
}

\newenvironment{SentenceExample}[1][]{
\setcounter{SentenceCounter}{0}
\vspace{0.5mm}
\begin{table}[htb]
\centering
\bgroup

\begin{tabular}{|p{0.45\textwidth}|}
\hline
\footnotesize
\vspace{1mm}
\def\temp{#1}
\ifx\temp\empty
    \refstepcounter{ExampleCounter}
    \noindent\textbf{Example \theExampleCounter}
\else
    \noindent\textbf{Example \theExampleCounter\hspace{1mm}(#1)}
\fi
\small
\begin{itemize}[leftmargin=*]

}{
\end{itemize}
\\\hline
\end{tabular}
\egroup
\label{eg\theExampleCounter}
\end{table}
\vspace{0.5mm}
}

\newcommand{\eg}[1]{Example #1}

\newcommand{\sent}[2]{sentence~#1.#2}
    
\begin{document}

\title{Identifying Relationships Among Sentences in Court Case Transcripts using Discourse Relations}


\author{\IEEEauthorblockN{Gathika Ratnayaka, Thejan Rupasinghe, Nisansa de Silva, \\ Menuka Warushavithana, Viraj Gamage, Amal Shehan Perera}
\IEEEauthorblockA{Department of Computer Science \& Engineering\\University of Moratuwa}
\IEEEauthorblockA{Email: gathika.14@cse.mrt.ac.lk}
}

\maketitle

\begin{abstract}
Case Law has a significant impact on the proceedings of legal cases. Therefore, the information that can be obtained from previous court cases is valuable to lawyers and other legal officials when performing their duties. This paper describes a methodology of applying discourse relations between sentences when processing text documents related to the legal domain. In this study, we developed a mechanism to classify the relationships that can be observed among sentences in transcripts of United States court cases. First, we defined relationship types that can be observed between sentences in court case transcripts. Then we classified pairs of sentences according to the relationship type by combining a machine learning model and a rule-based approach. The results obtained through our system were evaluated using human judges. To the best of our knowledge, this is the first study where discourse relationships between sentences have been used to determine relationships among sentences in legal court case transcripts. 
\end{abstract}


\section{INTRODUCTION}

Case Law can be described as a part of common law, consisting of judgments given by higher (appellate) courts in interpreting the statutes (or the provisions of a constitution) applicable in cases brought before them \cite{caseLawDefinition}. In order to make use of the case law, lawyers and other legal officials have to manually go through related court cases to find relevant information. This task requires a significant amount of effort and time. Therefore, automatic extraction of Information from legal court case transcripts would generate numerous benefits to the people working in the legal domain. 

From this point onwards we are referring to the \textit{court case transcripts} as \textit{court cases}. In the process of extracting information from legal court cases, it is important to identify how arguments and facts are related to one another. The objective of this study is to automatically determine the relationships between sentences which can be found in documents related to previous court cases of United States Supreme Court. Transcripts of U.S. court cases were obtained from FindLaw\footnote{https://caselaw.findlaw.com/} following a method similar to numerous other artificial intelligence applications in the legal domain~\cite{jayawardana2017word,sugathadasa2017synergistic,jayawardana2017deriving,sugathadasa2018Legal,jayawardana2017semi}.

When a sentence in a court case is considered, it may provide details on arguments or facts related to a particular legal situation. Some sentences may elaborate on the details provided in the previous sentence. It is also possible that the following sentence may not have any relationship with the details in the previous sentence and may provide details about a completely new topic. Another type of relationship is observed when a sentence provides contradictory details to the details provided in the previous sentence. Determining these relationships among sentences is vital to identifying the information flow within a court case. To that end, it is important to consider the way in which clauses, phrases, and text are related to each other. It can be argued that identifying relationships between sentences would make the process of Information Extraction from court cases more systematic given that it will provide a better picture of the information flow of a particular court case. To achieve this objective, we used discourse relations based approach to determine the relationships between sentences in legal documents.

Several theories related to discourse structures have been proposed in recent years.  Cross-document Structure Theory (CST) \cite{radev2000common} , Penn Discourse Tree Bank (PDTB) \cite{rashmi2008penn}, Rhetorical Structure Theory (RST) \cite{mann1987rhetorical,carlson2002rst} and Discourse Graph Bank \cite{wolf2004discourse} can be considered as prominent discourse structures. The main difference that can be observed between each of these discourse structures is they have defined the relation types in a different manner. This is mainly due to the fact that different discourse structures are intended for different purposes. In this study, we have based the discourse structure on the discourse structure proposed by CST.

A sentence in a court case transcript can contain different types of details such as descriptions of a scenario, legal arguments, legal facts or legal conditions. The main objective of identifying relationships between sentences is to determine which sentences are connected together within a single flow. If there is a weak or no relation between two sentences, it would probably infer that those two sentences provide details on different topics. Consider the sentence pair taken from \textit{Lee v. United States}~\cite{1977lee} shown in~\eg{1}.

\begin{SentenceExample}
\item \begin{Sentence} The Government makes two errors in urging the adoption of a per se rule that a defendant with no viable defense cannot show prejudice from the denial of his right to trial.\end{Sentence}

\item \begin{Sentence} First, it forgets that categorical rules are ill suited to an inquiry that demands a "case-by-case examination" of the "totality of the evidence".\end{Sentence}
\end{SentenceExample}

It can be seen that \sent{1}{2} elaborates further on the details provided by \sent{1}{1} to give a more comprehensive idea on the topic which is discussed in \sent{1}{1}. These two sentences are connected to each other within a same flow of information. This can be considered as \textit{Elaboration} relationship, which is a relation type described in CST. Now, Consider the sentence pair shown in~\eg{2} which was also taken from \textit{Lee v. United States}~\cite{1977lee}.

\begin{SentenceExample}
\item \begin{Sentence}Courts should not upset a plea solely because of post hoc assertions from a defendant about how he would have pleaded but for his attorney's deficiencies.\end{Sentence}

\item \begin{Sentence}Rather, they should look to contemporaneous evidence to substantiate a defendant's expressed preferences.\end{Sentence}
\end{SentenceExample}

In~\eg{2}, it can be seen that the two sentences have the \textit{Follow Up} relationship as defined in CST. But still, these two sentences are connected together within the same information flow in a court case. There are also situations where we can see sentences are showing characteristics which are common to multiple discourse relations. Therefore, several discourse relations can be grouped together based on their properties to make the process of determining relationships between sentences in court case transcripts more systematic. 

The two sentences for \eg{3} were also taken from \textit{Lee v. United States} \cite{1977lee}:   

\begin{SentenceExample}
\item \begin{Sentence}The question is whether Lee can show he was prejudiced by that erroneous advice.\end{Sentence}

\item \begin{Sentence}A claim of ineffective assistance of counsel will often involve a claim of attorney error "during the course of a legal proceeding"--for example, that counsel failed to raise an objection at trial or to present an argument on appeal.\end{Sentence}
\end{SentenceExample}


The \sent{3}{2} follows \sent{3}{1}. A significant connection between these two sentences cannot be observed. It can also be seen that \sent{3}{2} starts a new flow by deviating from the topic discussed in \sent{3}{1}. These observations which were provided by analyzing court cases emphasize the importance of identifying relationships between sentences. 



In this study, we defined the relationship types that are important to be considered when it comes to information extraction from court cases. Next, for each of the relationship type we defined, we identified the relevant CST relations \cite{radev2000common}. Finally, we developed a system to predict the relationship between given two sentences of a court case transcript by combining a machine learning model and a rule-based component. 

The next section provides an overview of how discourse relations have been applied in different domains including the legal domain. Section III describes the methodology which was followed when developing our system. Section IV describes the approaches we took to evaluate the system. The results obtained by evaluating the system are analyzed in Section V. Finally, we conclude our discussion in Section VI.    

\section{Background}

Understanding how information is related to each other in machine-readable texts has always been a challenge when it comes to Natural Language Processing. Determining the way in which two textual units are connected to each other is helpful in different applications such as text classification, text summarization, understanding the context, evaluating answers provided for a question. Analyzing of discourse relationships or rhetorical relationships between sentences can be considered as an effective approach to understanding the way how two textual units are connected with each other.

Discourse relations have been applied in different application domains related to NLP.  \cite{zhang2002towards} describes CST\cite{radev2000common} based text summarization approach which involves mechanisms such as identifying and removing redundancy in a text by analyzing discourse relations among sentences. \cite{uzeda2009comprehensive} compares and evaluates different methods of text summarizations which are based on RST \cite{carlson2002rst}. In another study \cite{castro2010experiments}, text summarization has been carried out by ranking sentences based on the number of discourse relations existing between sentences.\cite{marcu1997discourse,radev2004centroid,louis2010discourse} are some other studies where discourse analysis has been used for text summarization. These studies related to text summarization suggest that discourse relationships are useful when it comes identifying information that discusses on same topic or entity and also to capture information redundancy. Analysis of discourse relations has also been used for question answering systems \cite{litkowski2001cl,verberne2007discourse} and for natural language generation \cite{piwek2010generating}.

In the study \cite{ExploitingDiscourseR}, discourse relations existing between sentences are used to generate clusters of similar sentences from document sets. This study shows that a pair of sentences can show properties of multiple relation types which are defined in CST \cite{radev2000common}. In order to facilitate text clustering process, discourse relations have been redefined in this study by categorizing overlapping or closely related CST relations together.  In \cite{zahri2015exploiting}, the discourse relationships which are defined in \cite{ExploitingDiscourseR} have been used for text summarization based on text clustering. The studies \cite{ExploitingDiscourseR,zahri2015exploiting} emphasize how discourse relationships can be defined according to the purpose and objective of the study in order to enhance the effectiveness.

When it comes to the legal domain, \cite{moens1999information} discusses the potential of discourse analysis for extracting information from legal texts. \cite{hachey2004rhetorical} describes a classifier which determines the rhetorical status of a sentence from a corpus of legal judgments. In this study, rhetorical annotation scheme is defined for legal judgments. The study \cite{hachey2006extractive} provides details on summarization of legal texts using rhetorical annotation schemes. The studies \cite{hachey2004rhetorical, hachey2006extractive} focus mainly on the rhetorical status in a sentence, but not on the relationships between sentences. An approach which can be used to detect the arguments in legal text using lexical, syntactic, semantic and discourse properties of the text is described in \cite{moens2007automatic}.   

In contrast to other studies, this study is intended to identify relationships among sentences in court case transcripts by analyzing discourse relationships between sentences. Identifying relationships among sentences will be useful in the task of determining how information is flowed within a court case.

\section{Methodology}

\subsection{Defining Discourse Relationships in Court Cases}
 
Five major relationship types were defined by examining the nature of relationships that can be observed between sentences in court case transcripts.   

\begin{itemize}

\item \textbf{Elaboration -} One sentence adds more details to the information provided in the preceding sentence or one sentence develops further on the topic discussed in the previous sentence.  
\item \textbf{Redundancy -} Two sentences provide the same information without any difference or additional information.
\item \textbf{Citation -} A sentence provides references relevant to the details provided in the previous sentence.
\item \textbf{Shift in View -} Two sentences are providing conflicting information or different opinions on the same topic or entity.
\item \textbf{No Relation -} No relationship can be observed between the two sentences. One sentence discusses a topic which is different from the topic discussed in another sentence.
\end{itemize}
After defining these relationships, we adopted the rhetorical relations provided by CST \cite{radev2000common} to align with our definitions as shown in the Table below.

\begin{table}[h]
\caption{Adopting CST Relationships}
\label{table_adopting_cst_relationships}
\begin{center}
\begin{tabular}{|l|l|}
\hline
\textbf{Definition} & \textbf{CST Relationships}\\
\hline
\hline
Elaboration & \begin{tabular}[c]{@{}l@{}}Paraphrase,Modality,Subsumption,Elaboration, \\ Indirect Speech,
Follow-up, Overlap,
\\Fulfillment, Description, Historical Background,
 \\ Reader Profile,Attribution
\end{tabular}\\
\hline
Redundancy & Identity\\
\hline
Citation & Citation\\
\hline
Shift in View & \begin{tabular}[c]{@{}l@{}}Change of Perspective,Contradiction \\ 
\end{tabular}\\
\hline
No Relation & -\\
\hline

\end{tabular}
\end{center}
\end{table}

It is very difficult to observe same sentence appearing more than once within nearby sentences in court case transcripts. However, we have included it as a relationship type in order to identify redundant information in a case where the two sentences in a sentence pair are the same. 

\subsection{Expanding the Dataset}

A Machine Learning model was developed in order to determine the relationship between two sentences in court cases. We used the publicly available dataset of CST bank \cite{Radev&al.03} to learn the Model. 
The dataset obtained from CST bank contains sentence pairs which are annotated according to the CST relation types. Since we have a labeled dataset \cite{Radev&al.03}, we performed supervised learning to develop the machine learning model.  Support Vector Machine (SVM) was used because it has shown promising results in previous studies where discourse relations have been used to identify relationships between sentences \cite{ExploitingDiscourseR,zahri2015exploiting}.

Table \ref{table_CSTBank} provides details on the number of sentence pairs in the data set for each relationship type.

\begin{table}[h]
\caption{Number of sentence pairs for each relationship type}
\label{table_CSTBank}
\begin{center}
\begin{tabular}{|l|c|}
\hline
\textbf{CST Relationship} & \textbf{Number of Sentence Pairs}\\ 
\hline
\hline
Identity & 99\\
\hline
Equivalent & 101\\
\hline
Subsumption & 590\\
\hline
Contradiction & 48\\ 
\hline
Historical Background & 245 \\
\hline 
Modality & 17\\
\hline
Attribution & 134\\
\hline
Summary & 11 \\
\hline
Follow-up & 159 \\
\hline
Indirect Speech & 4 \\
\hline
Elaboration & 305 \\
\hline
Fulfillment & 10\\
\hline
Description & 244\\
\hline
Overlap (Partial Equivalence) & 429 \\
\hline
\end{tabular}
\end{center}
\end{table}

By examining the CST relationship types available in the dataset Table\ref{table_CSTBank}, it can be observed that a relationship type which suggests that there is no relationship between sentences cannot be found. But \textit{No Relation} is a fundamental relation type that can be observed between two sentences in court case transcripts. Therefore, we expanded the data set by manually annotating 50 pairs of sentences where a relationship between two sentences cannot be found. This new class was named as \textit{No Relation}. The 50 sentence pairs which were annotated were obtained from previous court case transcripts.

A sentence pair is made up of a source sentence and a target sentence. The source sentence is compared with the target sentence when determining the relationship that is present in the sentence pair. For example, if the source sentence contains all the information in target sentence with some additional information, the sentence pair is said to have the subsumption relationship. Similarly, if the source sentence elaborates the target sentence, the sentence pair is said to have the elaboration relationship. 

\subsection{Determining the relationship between sentences using SVM Model}

In order to train the SVM model with annotated data, features based on the properties that can be observed in a pair of sentences were defined. Before calculating the features related to words, we removed \textit{stop words} in sentences to eliminate the effect of less significant words. Also, co-referencing was performed on a given pair of sentences using Stanford CoreNLP CorefAnnotator (\textit{coref}) \cite{clark2015entity} in order to make the feature calculation more effective. The two sentences for \eg{4} are also taken from \textit{Lee v. United States} \cite{1977lee},

\begin{SentenceExample}
\item \begin{Sentence}[Target]Petitioner Jae Lee moved to the United States from South Korea with his parents when he was 13.\end{Sentence} 

\item \begin{Sentence}[Source]In the 35 years he has spent in this country, he has never returned to South Korea, nor has he become a U. S. citizen, living instead as a lawful permanent resident.\end{Sentence} 
\end{SentenceExample}

Here the ``Petitioner Jae Lee" in the target sentence, is referred using the pronouns ``he" and ``his" in both sentences. the system replaces “he” and “his” with their \textit{representative mention} “Petitioner Jae Lee”. Then the sentences in Example 4 are changed as shown below.;

\begin{SentenceExample}[updated]
\item \begin{Sentence}[Target]Petitioner Jae Lee moved to the United States from South Korea with \textbf{Petitioner Jae Lee} parents when \textbf{Petitioner Jae Lee} was 13.\end{Sentence}

\item \begin{Sentence}[Source]In the 35 years \textbf{Petitioner Jae Lee} has spent in this country, \textbf{Petitioner Jae Lee} has never returned to South Korea, nor has \textbf{Petitioner Jae Lee} become a U. S. citizen, living instead as a lawful permanent resident.\end{Sentence}
\end{SentenceExample}

By resolving co-references calculating Noun Similarity, Verb Similarity, Adjective Similarity, Subject Overlap Ratio, Object Overlap Ratio, Subject Noun Overlap Ratio and Semantic Similarity between Sentence features were made more effective.

All the features were calculated and normalized such that their values fall into $[0,1]$ range. We have defined 9 feature categories based on the properties that can be observed in a pair of sentences.

Following 5 feature categories were adopted mainly from  \cite{ExploitingDiscourseR} though we have done changes in implementation such as use of co-referencing.\\


\newcounter{para}
\newcommand\feature{\par\refstepcounter{para}\noindent\thepara.\space}

\feature Cosine Similarities

Following cosine similarity values are calculated for a given sentence pairs,
		\begin{itemize}
		\item Word Similarity
        \item Noun Similarity
        \item Verb Similarity
        \item Adjective Similarity
		\end{itemize}
        
Following equation is used to calculate the above mentioned cosine similarities.

\begin{equation} \label{eq:1}
CosineSimilarity = \frac{\sum_{i=1}^{n} FV_{S,i} * FV_{T,i}}{\sqrt[2]{\sum_{i=1}^{n} (FV_{S,i})^{2}}+\sqrt[2]{\sum_{i=1}^{n} (FV_{T,i})^{2}}}
\end{equation}

Here $FV_{S,i}$ and $FV_{T,i}$ represents frequency vectors of source sentence and target sentence respectively.  Standford CoreNLP POS Tagger (\textit{pos}) \cite{toutanova2003feature} is used to identify nouns, verbs and adjectives in sentences.

In calculating the Noun Similarity feature, singular and plural nouns, proper nouns, personal pronouns and possessive pronouns are considered. Both superlative and comparative adjectives are considered when calculating the Adjective Similarity. The system ignores verbs that are lemmatized into \textit{be}, \textit{do}, \textit{has} verbs when calculating Verb Similarity feature as the priority should be given to effective verbs in sentences.
\linebreak

\feature Word Overlap Ratios

Two ratios are considered based on the word overlapping. One ratio is measured in relation to the target sentence. Another ratio is measured in relation to the source sentence. These ratios provide an indication on the equivalence of two sentences. For example, when it comes to a relationship like subsumption, source sentence usually contains all the words in the target sentence. This property will be also useful in determining relations such as Identity, Overlap (Partial Equivalence) which are based on the equivalence of two sentences.

\begin{equation} \label{eq:2}
WOR(T)=\frac{Comm(T,S)}{Distinct(T)}
\end{equation}

\begin{equation} \label{eq:3}
WOR(S) = \frac{Comm(T,S)}{Distinct(S)}
\end{equation}

$WOR(T)$, $WOR(S)$ represents the word overlap ratios measured in relation to source and target sentences respectively. $Distinct(T)$, $Distinct(S)$ represents number of distinct words in source sentence and target sentence respectively. The number of distinct common words between two sentences are shown by $Comm(T,S)$.
\linebreak

\feature Grammatical Relationship Overlap Ratios

Three ratios which represent the grammatical relationship between target and source sentences are considered.

\begin{itemize}
\item Subject Overlap Ratio
\begin{equation} \label{eq:4}
SubjOverlap = \frac{Comm(Subj(S), Subj(T))}{Subj(S)}
\end{equation}

\item Object Overlap Ratio
\begin{equation} \label{eq:5}
ObjOverlap = \frac{Comm(Obj(S), Obj(T))}{Obj(S)}
\end{equation}

\item Subject Noun Overlap Ratio
\begin{equation} \label{eq:6}
SubjNounOverlap = \frac{Comm(Subj(S), Noun(T))}{Subj(S)}
\end{equation}
\end{itemize}

All these features are calculated with respect to the source sentence. $Subj, Obj, Noun$ represents the number of subjects, objects, and nouns respectively. $Comm$ gives the number of common elements.

Stanford CoreNLP Dependency Parse Annotator (\textit{depparse}) \cite{chen2014fast} is used here to identify subjects and objects. All the subject types including nominal subject, clausal subject, their passive forms and controlling subjects are taken into account in calculating the number of subjects. Direct and indirect objects are considered when calculating the number of objects. All subject and object types are referred from \textit{Stanford typed dependencies manual}~\cite{de2008stanford}.
\linebreak

\feature Longest Common Substring Ratio

Longest Common Substring is the maximum length word sequence which is common to both sentences. When the number of characters in longest common substring is taken as $n(LCS)$ and number of characters in source sentence is taken as $n(S)$, Longest Common Substring Ratio ($LCSR$) can be calculated as,

\begin{equation} \label{eq:7}
LCSR=\frac{n(LCS)}{n(S)}
\end{equation}

This value indicates the part of the target sentence which is present in the source sentence as a fraction. Thus, this will be useful especially in determining discourse relations such as overlap, attribution, and paraphrase.
\linebreak
 
\feature Number of Entities

Ratio between number of named entities can be used as a measurement of relationship between two sentences.

\begin{equation} \label{eq:8}
NERatio=\frac{NE(S)}{Max(NE(S), NE(T))}
\end{equation}

$NE(X)$ represents the number of named entities in a given sentence $X$. Stanford CoreNLP Named Entity Recognizer (NER) \cite{finkel2005incorporating} was used to identify named entities which belong to 7 types; PERSON, ORGANIZATION, LOCATION, MONEY, PERCENT, DATE and TIME.
\linebreak

In addition to the above mentioned features, following features have been introduced to the system.
\linebreak

\setcounter{para}{0}

\feature Semantic Similarity between Sentences
 
 This feature is useful in determining the closeness between two sentences. Semantic similarity will provide the closeness between those two words. A method described in \cite{tayal2014word} is adopted when calculating the semantic similarity between two sentences. Semantic similarity score for a pair of sentences is calculated using WordNet::Similarity \cite{pedersen2004wordnet}. 
 
\begin{equation} \label{eq:9}
score = Average\Bigg(\sum_{i=1}^{n} NounScore+\sum_{i=1}^{n} VerbScore\Bigg)
\end{equation}
\linebreak

\feature Transition Words and Phrases
 
 Availability of a transition word or a transition phrase at the start of a sentence indicates that there is a high probability of having a strong relationship with the previous sentence. For example, sentences beginning with transition words such as \textit{And}, \textit{Thus} usually elaborates the previous sentence. Phrases such as \textit{To make that}, \textit{In addition} at the beginning of a sentence also implies that the sentence is elaborating on the details provided in the previous sentence. Considering these linguistic properties two boolean features were defined.

\begin{enumerate}
    \item Elaboration Transition: If the first word of the source sentence is a transition word which implies elaboration such as \textit{and}, \textit{thus}, \textit{therefore} or if a transition phrase is found within first six words of the source sentence, this feature will output 1. If both of above two conditions are false, the feature will return 0. We maintain two lists containing 59 transition words and 91 transition phrases which implies elaboration. Though it is difficult to include all transition phrases in the English language which implies elaboration relationship, we can clearly say that if these phrases are present at the beginning of a sentence, the sentence is more than likely to elaborate the previous sentence.

    \item Follow Up Transition: If the source sentence begins with a word like \textit{however}, \textit{although} or phrases like \textit{in contrast}, \textit{on the contrary} which implies that the source sentence is following up the target sentence, this feature will output 1. Otherwise, the feature will output 0.\\ 
\end{enumerate}


\feature Length Difference Ratio
 
 This feature considers the difference of lengths between the source sentence and the target sentence. When $length(S)$ and $length(T)$ represent the number of words in source sentence and target sentence respectively, Length Difference Ratio ($LDR$) is calculated as shown below.
 
\begin{equation} \label{eq:10}
LDR = 0.5 + \frac{length(S) - length(T)}{2 * Max(length(S),length(T))}
\end{equation}
 
 In a relationship like Subsumption, the length of the source sentence has to be more than the length of the target sentence. In Identity relationship, both sentences are usually of the same length. These properties can be identified using this feature. 
 \linebreak
 
\feature Attribution

This feature checks whether a sentence describes a detail in another sentence in a more descriptive manner. Within this feature, we check whether a word or phrase in one sentence is cited in the other sentence using a quotation mark to determine this property. This is also a boolean feature. The source sentence and target sentence for \eg{5} was obtained from \textit{Turner v. United States} \cite{1970turner}:

\begin{SentenceExample}
\item \begin{Sentence}[Target]Such evidence is 'material' . . . when there is a reasonable probability that, had the evidence been disclosed, the result of the proceeding would have been different.\end{Sentence}

\item \begin{Sentence}[Source]A 'reasonable probability' of a different result is one in which the suppressed evidence 'undermines confidence in the outcome of the trial.\end{Sentence}
\end{SentenceExample}

It can be seen that source sentence define or provides more details on what is meant by "reasonable probability" in the target sentence. Such properties can be identified using this feature.

\subsection{Determining Explicit Citation Relationships in Court Case Transcripts}
        
In legal court case documents, several standard ways are used to point out whence a particular fact or condition was obtained. The target sentence and source sentence in \eg{6} are obtained from  \textit{Lee v. United States}\cite{1977lee}. 
        
\begin{SentenceExample}
\item \begin{Sentence}[Target]The decision whether to plead guilty also involves assessing the respective consequences of a conviction after trial and by plea.\end{Sentence}

\item \begin{Sentence}[Source]See INS v. St. Cyr, 533 U. S. 289, 322-323 (2001).\end{Sentence}
\end{SentenceExample}

The two sentences given in \eg{6} are adjacent to each other. It can be clearly seen that the source sentence provides a citation for the target sentence. This is only one of the many ways of providing citations in court case transcripts.  

After observing different ways of providing citations in court case transcripts, a rule-based mechanism to detect such citations was developed. If this rule-based system detects that there is citation relationship, the pair of sentences will be assigned with the citation relationship. Such a pair of sentences will not be inputted to the SVM model for further processing.

\section{Experiments}
In order to determine the effectiveness of our system, it is important to carry out evaluations using legal court case transcripts, as it is the domain this system is intended to be used. Court case transcripts related to United States Supreme Court were obtained from Findlaw. Then the transcripts were preprocessed in order to remove unnecessary data and text. Court case title, section titles are some examples of details which were removed in the preprocessing process. Those details are irrelevant when it comes to determining relationships between sentences.
        
The relationship types of sentence pairs were assigned using the system. First, the pairs were checked for citation relationship using the rule-based approach. The relationship types of the sentence pairs where citation relationship couldn't be detected using the rule-based approach were determined using the Support Vector Machine model.
        
The results obtained using the system for the sentence pairs extracted from the court case transcripts were then stored in a database. From those sentence pairs, 200 sentence pairs were selected to be annotated by human judges. Before selecting 200 sentence pairs, the sentence pairs were shuffled to eliminate the potential bias that could have been existent due to a particular court case. Shuffling was helpful in making sure that the sentence pairs to be annotated by human judges were related to different court case transcripts.
        
Then the selected 200 pairs of sentences to be annotated were grouped together as clusters of five sentence pairs. Each cluster was annotated by two human judges who were trained to identify the relationships between sentence pairs as defined in this study.

\section{Results}

As expected, the redundancy relationships between sentences could not be observed within the sentence pairs which were annotated using human judges. From the 200 sentence pairs that were observed, our system did not predict redundancy relationship for any sentence pair. Similarly, human judges did not annotate the "redundancy" relationship for any sentence pair.

The confusion matrix which was generated according to the results obtained is given in Table \ref{table_confusion_matrix}. The details provided in the matrix are based only on the sentence pairs that were agreed by two human judges to have a similar relationship. The reasoning behind this approach is to eliminate sentence pairs where there are ambiguities of the relationship type between them. 

The same approach was used to obtain the results which are presented in Table \ref{table_results_comparison_both_agree}. In contrast, Table \ref{table_results_comparison_at_least_one_judge} contains results obtained by considering sentence pairs where at least one of the two judges who annotated the pair agrees upon a particular relationship type.

\newcommand{\shadedCell}[1]{
#1\% 
\cellcolor{gray!#1!white}
}

\begin{table}[h]
\caption{confusion matrix}
\label{table_confusion_matrix}
\begin{center}
\begin{tabular}{|l|c|c|c|c|c|}\hline
 \diagbox[width=8em]{Actual}{Predicted}& \rot{Elaboration} & \rot{No Relation} & \rot{Citation} & \rot{Shift In View} & $\Sigma$ \\
 \hline
 Elaboration& \shadedCell{93.9} & \shadedCell{6.1} & \shadedCell{0.0} & \shadedCell{0.0} & 99\\
 \hline
 No Relation& \shadedCell{11.9} & \shadedCell{88.1} & \shadedCell{0.0} & \shadedCell{0.0} & 42\\
 \hline
 Citation & \shadedCell{0.0} & \shadedCell{4.8} & \shadedCell{95.2} & \shadedCell{0.0} & 21\\
 \hline
 Shift In View& \shadedCell{100.0} & \shadedCell{0.0} & \shadedCell{0.0} & \shadedCell{0.0} & 3\\
 \hline
$\Sigma$ & 101 & 44 & 20 & 0 & 165\\
 \hline
\end{tabular}
\end{center}
\end{table}

\begin{table}[h]
\caption{results comparison of pairs where both judges agree}
\label{table_results_comparison_both_agree} 
\begin{center}
\begin{tabular}{|c|c|c|c|}
\hline
Discourse Class & Precision & Recall & F-Measure \\
\hline
Elaboration & 0.921 & 0.939 & 0.930\\
\hline
No Relation & 0.841 & 0.881 & 0.861\\
\hline
Citation & 1.000 & 0.952 & 0.975\\
\hline
Shift in View & - & 0 & -\\
\hline

\end{tabular}
\end{center}
\end{table}

\begin{table}[h]
\caption{results comparison of pairs where at least one judge agrees}
\label{table_results_comparison_at_least_one_judge} 
\begin{center}
\begin{tabular}{|c|c|c|c|}
\hline
Discourse Class & Precision & Recall & F-Measure \\
\hline
Elaboration & 0.930 & 0.902 & 0.916 \\
\hline
No Relation & 0.846 & 0.677 & 0.752 \\
\hline
Citation & 1.000 & 0.910 & 0.953 \\
\hline
Shift in View & - & 0 & - \\
\hline

\end{tabular}
\end{center}
\end{table}

 The \textit{Recall} results given in Table \ref{table_results_comparison_both_agree} has a significant importance as all the sentence pairs contained in that results set are annotated with a relationship type which was agreed by two human judges. The \textit{Precision} results provided in Table \ref{table_results_comparison_at_least_one_judge} indicate the probability of at least one human judge agreeing with the system's predictions in relation to each relationship type. Evaluation results from Table \ref{table_results_comparison_both_agree}, Table \ref{table_results_comparison_at_least_one_judge} the system works well when identifying \textit{Elaboration}, \textit{No Relation} and \textit{Citation} relationship types where F-measure values are above 75\% in all cases. \textit{Shift in View} relationship type was not assigned by the system to any of the 200 sentence pairs which were considered in the evaluation. 

Human vs Human correlation and Human vs System correlation when it comes to identifying these relationship types were also analyzed. First, we calculated these correlations without considering the relationship type using the following approach. For a given sentence pair $P$, $m(P)$ is the value assigned to the pair. $n$ is the number of sentence pairs. 
\linebreak

\setcounter{para}{0}

\feature Human vs Human Correlation ($Cor(H,H)$)

When both human judges are agreeing on a single relationship type for the pair $P$, we assign $m(P) = 1$. Otherwise, we assign $m(P)=0$.

\begin{equation}\label{eq:11}
Cor(H,H) = \frac{\sum_{P=1}^{n} m(P)}{n}
\end{equation}

\feature Human vs System Correlation ($Cor(H,S)$)

When both human judges are agreeing with the relationship type predicted by the system for the sentence pair $P$, we assign $m(P) = 1.0$. If only one human judge is agreeing with the relationship type predicted by the system for $P$, we assign $m(P) = 0.5$. If both human judges disagree with the relationship type predicted by the system for $P$, we assign $m(P) = 0.0$. 

\begin{equation} \label{eq:12}
Cor(H,S) = \frac{\sum_{P=1}^{n} m(P)}{n}
\end{equation}

It was observed that the correlation between a human judge and another human judge was calculated to be $0.805$ while the correlation between a human judge and the system was calculated to be $0.813$.  
When analyzing these two correlations, it can be seen that our system performs with a capability which is close to the human capability.

The results obtained by calculating Human vs. Human and Human vs. System correlations in relation to each relationship type are given in Table \ref{table_6}. The results were obtained by Equation~\ref{eq:13} which calculates Human vs Human correlation ($Corr(H,H)$) and Equation~\ref{eq:14} which calculates Human vs System correlation ($Corr(H,S)$) where; for a given set $A$, $n(A)$ indicates number of elements in set $A$, and for a  relationship type $R$, $S$ denotes the set containing all the sentence pairs which are predicted by the system as having the relationship type $R$, $U$ denotes the set containing all the sentence pairs which were annotated by at least one human judge as having the relationship type $R$ and $V$ denotes the set containing all the sentence pairs which were annotated by two human judges as having the relationship type $R$.   







\begin{equation} \label{eq:13}
Corr(H,H) = \frac{n(V)}{n(U)}
\end{equation}

\begin{equation} \label{eq:14}
Corr(H,S) = \frac{n(S\wedge U)}{n(S \vee U)}
\end{equation}

The results obtained using this approach is provided in  Table \ref{table_6}. 

\begin{table}[h]
\caption{correlations by type}
\label{table_6}
\begin{center}
\begin{tabular}{|c|c|c|c|}
\hline
Discourse Class & Human-Human & Human-System& \begin{tabular}[c]{@{}l@{}} \underline{Human-System}
\\Human-Human\\ 
\end{tabular}\\
\hline
Elaboration & 0.75 & 0.843 & 1.124 \\
\hline
No Relation & 0.646 & 0.603 &  0.933 \\
\hline
Citation & 1.0 & 0.955  & 0.955 \\
\hline
Shift in View & 0.188 & 0.0 & 0.0\\
\hline

\end{tabular}
\end{center}
\end{table}

The results which are in Table \ref{table_6} suggest that the system performs with a capability which is near to the human capability when it comes to identifying relationships such as \textit{Elaboration}, \textit{No Relation}, and \textit{Citation} in court case transcripts. Enhancing system's ability to identify \textit{Shift in View} relationship is one of the major future challenges. At the same time Human vs Human correlation when it comes to identifying \textit{Shift in View} relationship type is 0.188. This indicates that humans are also having ambiguities when identifying \textit{Shift in View} relationships between sentences in court case transcripts.

Either \textit{Elaboration} or \textit{Shift in View} relationship occurs when the two sentences are discussing the same topic or entity. \textit{Shift in View} relationship occurs over \textit{Elaboration} when two sentences are providing different views or conflicting facts on the same topic or entity. The \textit{No Relation} relationship can be observed between two sentences when two sentences are no longer discussing the same topic or entity. In other words, \textit{No Relation} relationship suggests that there is a shift in the information flow in court cases. As shown in Table \ref{table_confusion_matrix}, the sentence pairs with \textit{Shift in View} relationship are always predicted as having \textit{Elaboration} relationship by the system. By observing these results, it can be seen that in most of the cases the system is able to identify whether the sentences are discussing the same topic or not.

\section{CONCLUSIONS}

The primary research contribution of this study was the use of discourse relationships between sentences to identify the relationships among sentences in transcripts of United States court cases. Five discourse relationship types were defined in this study in order to automatically identify the flow of information within a court case transcript. This study describes how a machine learning model and a rule-based system can be combined together to enhance the accuracy of identifying relationships between sentences in court case transcripts. Features based on the properties that can be observed between sentences have been introduced to enhance the accuracy of the machine learning model.

The proposed methodology can be successfully applied to identify the sentences which develop on the same discussion topic or entity. In addition, the system is capable of identifying situations in court cases where the discussion topic changes. The system is highly successful in the identification of legal citations. These outcomes demonstrate that the approach described in this study has a promising potential to be applied in tasks related to systematic information extraction from court case transcripts.One such task is the identification of supporting facts, citations which are related to a particular legal argument. Another is the identification of changes in discussion topics within a court case. 

The system has difficulties in detecting the occasions where the two sentences are providing different opinions on the same discussion topic. Enhancing this capability in the system can be considered as the major future work.


\bibliographystyle{IEEEtran}
\bibliography{references}

\begin{thebibliography}{10}
\providecommand{\url}[1]{#1}
\csname url@samestyle\endcsname
\providecommand{\newblock}{\relax}
\providecommand{\bibinfo}[2]{#2}
\providecommand{\BIBentrySTDinterwordspacing}{\spaceskip=0pt\relax}
\providecommand{\BIBentryALTinterwordstretchfactor}{4}
\providecommand{\BIBentryALTinterwordspacing}{\spaceskip=\fontdimen2\font plus
\BIBentryALTinterwordstretchfactor\fontdimen3\font minus
  \fontdimen4\font\relax}
\providecommand{\BIBforeignlanguage}[2]{{%
\expandafter\ifx\csname l@#1\endcsname\relax
\typeout{** WARNING: IEEEtran.bst: No hyphenation pattern has been}%
\typeout{** loaded for the language `#1'. Using the pattern for}%
\typeout{** the default language instead.}%
\else
\language=\csname l@#1\endcsname
\fi
#2}}
\providecommand{\BIBdecl}{\relax}
\BIBdecl

\bibitem{caseLawDefinition}
{WebFinance Inc}, ``What is case law? definition and meaning -
  businessdictionary.com,''
  \url{http://www.businessdictionary.com/definition/case-law.html}, (Accessed
  on 05/17/2018).

\bibitem{jayawardana2017word}
V.~Jayawardana, D.~Lakmal, N.~de~Silva, A.~S. Perera, K.~Sugathadasa,
  B.~Ayesha, and M.~Perera, ``Word vector embeddings and domain specific
  semantic based semi-supervised ontology instance population,''
  \emph{International Journal on Advances in ICT for Emerging Regions},
  vol.~10, no.~1, p.~1, 2017.

\bibitem{sugathadasa2017synergistic}
K.~Sugathadasa, B.~Ayesha, N.~de~Silva, A.~S. Perera, V.~Jayawardana,
  D.~Lakmal, and M.~Perera, ``Synergistic union of word2vec and lexicon for
  domain specific semantic similarity,'' in \emph{Industrial and Information
  Systems (ICIIS), 2017 IEEE International Conference on}.\hskip 1em plus 0.5em
  minus 0.4em\relax IEEE, 2017, pp. 1--6.

\bibitem{jayawardana2017deriving}
V.~Jayawardana, D.~Lakmal, N.~de~Silva, A.~S. Perera, K.~Sugathadasa, and
  B.~Ayesha, ``Deriving a representative vector for ontology classes with
  instance word vector embeddings,'' in \emph{Innovative Computing Technology
  (INTECH), 2017 Seventh International Conference on}.\hskip 1em plus 0.5em
  minus 0.4em\relax IEEE, 2017, pp. 79--84.

\bibitem{sugathadasa2018Legal}
K.~Sugathadasa, B.~Ayesha, N.~de~Silva, A.~S. Perera, V.~Jayawardana,
  D.~Lakmal, and M.~Perera, ``Legal document retrieval using document vector
  embeddings and deep learning,'' \emph{arXiv preprint arXiv:1805.10685}, 2018.

\bibitem{jayawardana2017semi}
V.~Jayawardana, D.~Lakmal, N.~de~Silva, A.~S. Perera, K.~Sugathadasa,
  B.~Ayesha, and M.~Perera, ``Semi-supervised instance population of an
  ontology using word vector embedding,'' in \emph{Advances in ICT for Emerging
  Regions (ICTer), 2017 Seventeenth International Conference on}.\hskip 1em
  plus 0.5em minus 0.4em\relax IEEE, 2017, pp. 1--7.

\bibitem{radev2000common}
D.~R. Radev, ``A common theory of information fusion from multiple text sources
  step one: cross-document structure,'' in \emph{Proceedings of the 1st SIGdial
  workshop on Discourse and dialogue-Volume 10}.\hskip 1em plus 0.5em minus
  0.4em\relax Association for Computational Linguistics, 2000, pp. 74--83.

\bibitem{rashmi2008penn}
P.~Rashmi, D.~Nihkil, L.~Alan, M.~Eleni, R.~Livio, J.~Aravind, W.~Bonnie
  \emph{et~al.}, ``The penn discourse treebank 2.0,'' in \emph{Proceedings of
  the Sixth International Conference on Language Resources and Evaluation
  (LREC’08), Marrakech, Morocco, may. European Language Resources Association
  (ELRA).}, 2008.

\bibitem{mann1987rhetorical}
W.~C. Mann and S.~A. Thompson, \emph{Rhetorical structure theory: A theory of
  text organization}.\hskip 1em plus 0.5em minus 0.4em\relax University of
  Southern California, Information Sciences Institute, 1987.

\bibitem{carlson2002rst}
L.~Carlson, M.~E. Okurowski, and D.~Marcu, \emph{RST discourse treebank}.\hskip
  1em plus 0.5em minus 0.4em\relax Linguistic Data Consortium, University of
  Pennsylvania, 2002.

\bibitem{wolf2004discourse}
F.~Wolf, E.~Gibson, A.~Fisher, and M.~Knight, ``Discourse graphbank,''
  \emph{Linguistic Data Consortium, Philadelphia}, 2004.

\bibitem{1977lee}
``{Lee v. United States},'' in \emph{US}, vol. 432, no. No. 76-5187.\hskip 1em
  plus 0.5em minus 0.4em\relax Supreme Court, 1977, p.~23.

\bibitem{zhang2002towards}
Z.~Zhang, S.~Blair-Goldensohn, and D.~R. Radev, ``Towards cst-enhanced
  summarization,'' in \emph{AAAI/IAAI}, 2002, pp. 439--446.

\bibitem{uzeda2009comprehensive}
V.~Uz{\^e}da, T.~Pardo, and M.~Nunes, ``A comprehensive summary informativeness
  evaluation for rst-based summarization methods,'' \emph{International Journal
  of Computer Information Systems and Industrial Management Applications
  (IJCISIM) ISSN}, pp. 2150--7988, 2009.

\bibitem{castro2010experiments}
M.~L. d.~R. Castro~Jorge and T.~A.~S. Pardo, ``Experiments with cst-based
  multidocument summarization,'' in \emph{Proceedings of the 2010 Workshop on
  Graph-based Methods for Natural Language Processing}.\hskip 1em plus 0.5em
  minus 0.4em\relax Association for Computational Linguistics, 2010, pp.
  74--82.

\bibitem{marcu1997discourse}
D.~Marcu, ``From discourse structures to text summaries,'' \emph{Intelligent
  Scalable Text Summarization}, 1997.

\bibitem{radev2004centroid}
D.~R. Radev, H.~Jing, M.~Sty{\'s}, and D.~Tam, ``Centroid-based summarization
  of multiple documents,'' \emph{Information Processing \& Management},
  vol.~40, no.~6, pp. 919--938, 2004.

\bibitem{louis2010discourse}
A.~Louis, A.~Joshi, and A.~Nenkova, ``Discourse indicators for content
  selection in summarization,'' in \emph{Proceedings of the 11th Annual Meeting
  of the Special Interest Group on Discourse and Dialogue}.\hskip 1em plus
  0.5em minus 0.4em\relax Association for Computational Linguistics, 2010, pp.
  147--156.

\bibitem{litkowski2001cl}
K.~C. Litkowski, ``Cl research experiments in trec-10 question answering,'' no.
  250.\hskip 1em plus 0.5em minus 0.4em\relax National Institute of Standards
  \& Technology, 2002, pp. 122--131.

\bibitem{verberne2007discourse}
S.~Verberne, L.~W.~J. Boves, N.~H.~J. Oostdijk, and P.~A. J.~M. Coppen,
  ``Discourse-based answering of why-questions,'' \emph{Traitement Automatique
  des Langues}, vol.~47, pp. 21--41, 2007.

\bibitem{piwek2010generating}
P.~Piwek and S.~Stoyanchev, ``Generating expository dialogue from monologue:
  motivation, corpus and preliminary rules,'' in \emph{Human Language
  Technologies: The 2010 Annual Conference of the North American Chapter of the
  Association for Computational Linguistics}.\hskip 1em plus 0.5em minus
  0.4em\relax Association for Computational Linguistics, 2010, pp. 333--336.

\bibitem{ExploitingDiscourseR}
N.~A.~H. Zahri, F.~Fukumoto, and S.~Matsuyoshi, ``Exploiting discourse
  relations between sentences for text clustering,'' in \emph{24th
  International Conference on Computational Linguistics}, 2012, p.~17.

\bibitem{zahri2015exploiting}
N.~A.~H. Zahri, F.~Fukumoto, M.~Suguru, and O.~B. Lynn, ``Exploiting rhetorical
  relations to multiple documents text summarization,'' \emph{International
  Journal of Network Security \& Its Applications}, vol.~7, no.~2, p.~1, 2015.

\bibitem{moens1999information}
M.-F. Moens, C.~Uyttendaele, and J.~Dumortier, ``Information extraction from
  legal texts: the potential of discourse analysis,'' \emph{International
  Journal of Human-Computer Studies}, vol.~51, no.~6, pp. 1155--1171, 1999.

\bibitem{hachey2004rhetorical}
B.~Hachey and C.~Grover, ``A rhetorical status classifier for legal text
  summarisation,'' \emph{Text Summarization Branches Out}, 2004.

\bibitem{hachey2006extractive}
------, ``Extractive summarisation of legal texts,'' \emph{Artificial
  Intelligence and Law}, vol.~14, no.~4, pp. 305--345, 2006.

\bibitem{moens2007automatic}
M.-F. Moens, E.~Boiy, R.~M. Palau, and C.~Reed, ``Automatic detection of
  arguments in legal texts,'' in \emph{Proceedings of the 11th international
  conference on Artificial intelligence and law}.\hskip 1em plus 0.5em minus
  0.4em\relax ACM, 2007, pp. 225--230.

\bibitem{Radev&al.03}
D.~Radev, J.~Otterbacher, and Z.~Zhang, ``{CSTBank: Cross-document Structure
  Theory Bank},'' http://tangra.si.umich.edu/clair/CSTBank, 2003.

\bibitem{clark2015entity}
K.~Clark and C.~D. Manning, ``Entity-centric coreference resolution with model
  stacking,'' in \emph{Proceedings of the 53rd Annual Meeting of the
  Association for Computational Linguistics and the 7th International Joint
  Conference on Natural Language Processing (Volume 1: Long Papers)}, vol.~1,
  2015, pp. 1405--1415.

\bibitem{toutanova2003feature}
K.~Toutanova, D.~Klein, C.~D. Manning, and Y.~Singer, ``Feature-rich
  part-of-speech tagging with a cyclic dependency network,'' in
  \emph{Proceedings of the 2003 Conference of the North American Chapter of the
  Association for Computational Linguistics on Human Language Technology-Volume
  1}.\hskip 1em plus 0.5em minus 0.4em\relax Association for Computational
  Linguistics, 2003, pp. 173--180.

\bibitem{chen2014fast}
D.~Chen and C.~Manning, ``A fast and accurate dependency parser using neural
  networks,'' in \emph{Proceedings of the 2014 conference on empirical methods
  in natural language processing (EMNLP)}, 2014, pp. 740--750.

\bibitem{de2008stanford}
M.-C. De~Marneffe and C.~D. Manning, ``Stanford typed dependencies manual,''
  Technical report, Stanford University, Tech. Rep., 2008.

\bibitem{finkel2005incorporating}
J.~R. Finkel, T.~Grenager, and C.~Manning, ``Incorporating non-local
  information into information extraction systems by gibbs sampling,'' in
  \emph{Proceedings of the 43rd annual meeting on association for computational
  linguistics}.\hskip 1em plus 0.5em minus 0.4em\relax Association for
  Computational Linguistics, 2005, pp. 363--370.

\bibitem{tayal2014word}
M.~A. Tayal, M.~Raghuwanshi, and L.~Malik, ``Word net based method for
  determining semantic sentence similarity through various word senses,'' in
  \emph{Proceedings of the 11th International Conference on Natural Language
  Processing}, 2014, pp. 139--145.

\bibitem{pedersen2004wordnet}
T.~Pedersen, S.~Patwardhan, and J.~Michelizzi, ``Wordnet:: Similarity:
  measuring the relatedness of concepts,'' in \emph{Demonstration papers at
  HLT-NAACL 2004}.\hskip 1em plus 0.5em minus 0.4em\relax Association for
  Computational Linguistics, 2004, pp. 38--41.

\bibitem{1970turner}
``{Turner v. United States},'' in \emph{US}, vol. 396, no. No. 190.\hskip 1em
  plus 0.5em minus 0.4em\relax Supreme Court, 1970, p. 398.

\end{thebibliography}

\end{document}